\documentclass{article}
\usepackage{spconf,amsmath,graphicx,hyperref}
\usepackage{booktabs}   % 提供 \toprule, \midrule, \bottomrule 等
\usepackage{multirow}   % 支持 \multirow
\usepackage{array}      % 更灵活的表格列格式
\usepackage{caption}    % 美化 caption（可选）
\usepackage{graphicx}   % 通常论文都会加载（可选）
\usepackage{amsmath, amssymb, amsfonts}
\usepackage{amsthm} % 用来定义定理、引理、定义等环境
\newtheorem{definition}{Definition}
\title{Recurrent Confidence Chain: Temporal-Aware Uncertainty Quantification  in Large Language Models}

\name{Zhenjiang Mao*, Anirudhh Venkat*\thanks{*First co-authors: equal contribution.}}
\address{University of Florida}
\begin{document}
%\ninept
%
\maketitle
\begin{abstract}
As reasoning modules, such as the chain-of-thought mechanism, are applied to large language models, they achieve strong performance on various tasks such as answering common-sense questions and solving math problems. The main challenge now is to assess the uncertainty of answers, which can help prevent misleading or serious hallucinations for users. Although current methods analyze long reasoning sequences by filtering unrelated tokens and examining potential connections between nearby tokens or sentences, the temporal spread of confidence is often overlooked. This oversight can lead to inflated overall confidence, even when earlier steps exhibit very low confidence. To address this issue, we propose a novel method that incorporates inter-step attention to analyze semantic correlations across steps. For handling long-horizon responses, we introduce a hidden confidence mechanism to retain historical confidence information, which is then combined with stepwise confidence to produce a more accurate overall estimate. We evaluate our method on the GAOKAO math benchmark and the CLadder causal reasoning dataset using mainstream open-source large language models. Our approach is shown to outperform state-of-the-art methods by achieving a superior balance between predictive quality and calibration, demonstrated by strong performance on both Negative Log-Likelihood and Expected Calibration Error.

\end{abstract}

\begin{figure*}[h!]
  \centering
  \includegraphics[width=\textwidth]{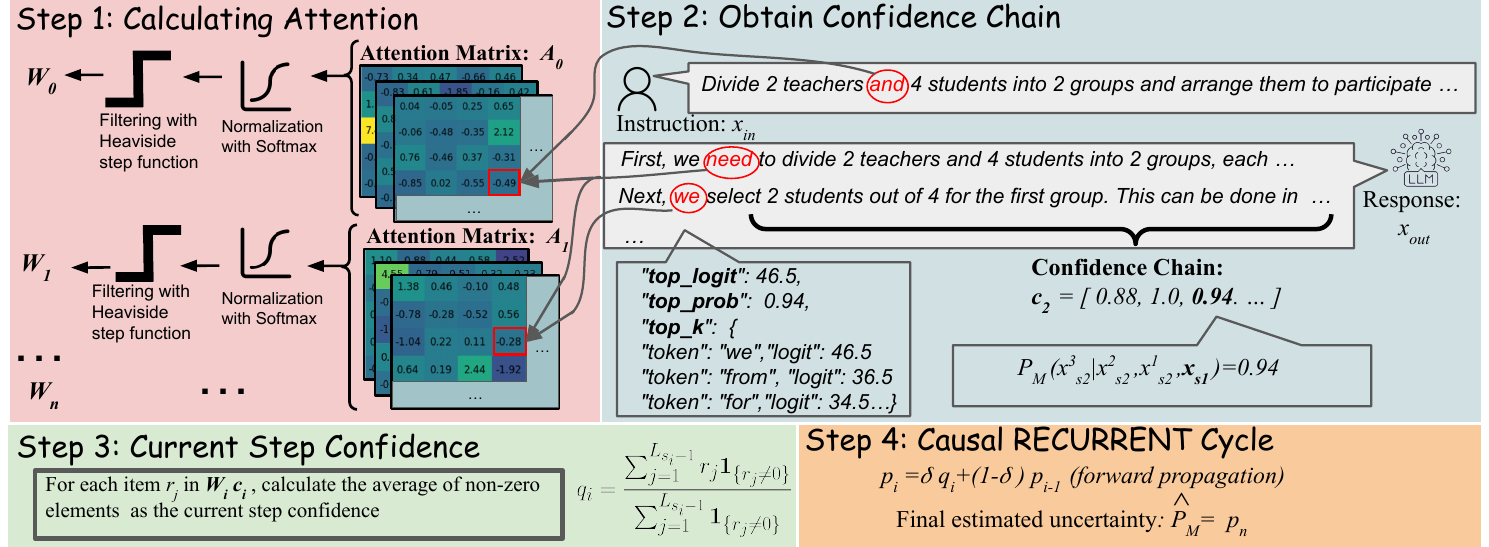}
  \caption{Overview of the Recurrent Confidence Chain (RCC) framework with four main steps.}
  \label{fig:overview}
\end{figure*}

\section{Introduction}
\label{sec:intro}

As Large Language Models (LLMs) are increasingly applied in daily life domains such as education and healthcare, their powerful abilities in deep reasoning can improve response generation, and the integration with external tools such as search engines can further enhance information verification. However, despite sophisticated reasoning workflow designs or the use of retrieval-augmented generation to add additional background and context, LLMs can still produce incorrect and unreliable responses, commonly referred to as factual errors and hallucinations. Thus, enabling LLMs to respond with \emph{``not sure"} instead of providing incorrect answers is an important way to give feedback to users when the models lack strong confidence. Uncertainty quantification for LLMs can provide a safeguard by notifying users when the model is not able to deliver a high-confidence answer.

Recent studies~\cite{shorinwa2025survey} adapt several approaches from traditional deep-learning methods to quantify uncertainty, such as self-verbalization, where LLMs generate numerical scores or uncertainty-related words to express their own confidence. Another major direction is self-consistency- or semantic-similarity-based methods, which require LLMs to generate multiple responses for the same instruction and assume that more consistent answers indicate higher confidence than those with significant variation. A further approach attempts to measure internal confidence by exploring mechanistic interpretability within the middle layers of LLMs~\cite{azaria2023the}, which typically requires additional fine-tuning and training and is therefore considered less convenient. Token-level uncertainty provides a white-box method by generating probabilities for each token~\cite{tian-etal-2023-just}, for example, using a softmax over the logit scores of the vocabulary, which requires no additional training and yields uncertainty simultaneously with inference.
% azaria2023internal vazhentsev2025uncertainty,tian2023just.ji-etal-2024-llm,

Despite the advancements, current research for token-level uncertainty quantification faces critical limitations:

\noindent
\textbf{Temporal confidence misalignment:} Current approaches treat reasoning sequences as flat probability distributions, failing to capture temporal dynamics of confidence propagation. Recent work~\cite{korbak2025chain} demonstrates that models exhibit high confidence in later steps despite early errors, creating inflated overall confidence. The problem worsens with auxiliary tokens (transition words, politeness markers) that dilute joint probability calculations~\cite{li2025language}.
%zhao2025chain

\noindent
\textbf{Semantic blindness between reasoning steps:} Existing methods~\cite{gao-etal-2024-spuq,zhang-etal-2024-luq} process tokens independently, ignoring semantic dependencies between consecutive steps. Recent analysis~\cite{da2025understanding} reveals that answer-level uncertainty measures cannot identify vulnerable reasoning paths. This leads to high confidence despite logical inconsistencies, particularly dangerous in mathematical and medical domains.
%gao2024spuq,liu2024uncertainty

\noindent
\textbf{Absence of confidence memory mechanisms:} Without confidence accumulation across extended chains, models cannot distinguish sustained uncertainty from isolated errors. While recent approaches~\cite{mao-etal-2025-temporalizing,duan2025uprop} attempt uncertainty propagation, they face exponential complexity. Empirical studies~\cite{he2025mmboundary} show that failing to assess step-wise confidence causes 7.5\% calibration degradation through cascading errors. Further evidence~\cite{liu2025adaptivestep} highlights this critical gap. To address these limitations, we propose a novel Recurrent Confidence Chain (RCC) framework with three key innovations:
%zhao2024saup,

\begin{itemize}
\item We introduce an inter-step attention mechanism to model semantic dependencies across reasoning steps, overcoming the semantic blindness of existing token-level methods.
% \item We introduce attention matrices between reasoning steps to model semantic dependencies, departing from independent token processing in existing work. Our threshold-based Heaviside filtering distinguishes meaningful connections from auxiliary content, directly addressing the semantic blindness problem in current approaches.
\item We design an efficient recurrent confidence propagation mechanism that maintains a history of uncertainty throughout the reasoning chain with linear complexity, avoiding the exponential growth of prior approaches.

% \item We develop a recurrent confidence propagation mechanism that maintains confidence history throughout reasoning chains. This solves accumulation problems while operating with linear complexity, avoiding the exponential growth that plagues existing methods.

\item We present a unified framework that integrates semantic correlation with temporal confidence tracking, enabling more robust uncertainty quantification for multi-step reasoning.

% \item We present a unified framework integrating attention-based correlation with temporal propagation. Our method seamlessly combines semantic dependency modeling with temporal confidence tracking for more reliable uncertainty quantification in multi-step reasoning.
\end{itemize}

Our experimental results on the GAOKAO math benchmark~\cite{zhang2023evaluating} and the CLadder causal reasoning dataset~\cite{jin2023cladder} demonstrate that RCC outperforms state-of-the-art methods, achieving a superior balance with both lower Negative Log-Likelihood (NLL) and Expected Calibration Error (ECE), thereby providing more trustworthy uncertainty estimates.

\begin{table*}[t]
\centering
\footnotesize
\caption{Performance Comparison of Uncertainty Quantification Methods on Reasoning Tasks.}
\label{tab:perf}
\setlength{\tabcolsep}{5pt}
\renewcommand{\arraystretch}{1.15}
\begin{tabular}{l *{8}{c}}
\toprule
\multirow{3}{*}{Model/Dataset/Methods}
  & \multicolumn{4}{c}{Qwen3-8B}
  & \multicolumn{4}{c}{Gemma3-12B} \\
\cmidrule(lr){2-5}\cmidrule(lr){6-9}
  & \multicolumn{2}{c}{MATH} & \multicolumn{2}{c}{CLadder}
  & \multicolumn{2}{c}{MATH} & \multicolumn{2}{c}{CLadder} \\
\cmidrule(lr){2-3}\cmidrule(lr){4-5}\cmidrule(lr){6-7}\cmidrule(lr){8-9}
  & NLL $\downarrow$ &ECE (\%) $\downarrow$ & NLL $\downarrow$ &ECE (\%) $\downarrow$ & NLL $\downarrow$ &ECE (\%) $\downarrow$ & NLL $\downarrow$ &ECE (\%) $\downarrow$
 \\
\midrule
Logits (Final)   & 4.369 & 18.95&5.722&21.21 &2.122 & 18.84&2.944&20.20
\\
Logits (Average) & 0.558 & 14.04 &0.552&12.57& 0.560 & 14.25 &0.494&11.31
\\
Self-Evaluation & 3.533 & 16.93 & 4.957&17.59 & 2.704 & 16.57 &1.340 &12.73
\\
Self-Consistency&1.573 &12.19 &3.763 & 11.42& 1.163&13.71 &2.109 &15.28
% \\
% Verbalized&
\\
SAR (Shifting Attention to Relevance)
&0.932&13.87&1.008&8.19&0.594&9.46&\textbf{0.465}&13.32\\
UQAC (Attention Chain) &0.729&10.07&2.271&6.36&0.503&13.35&0.933&\textbf{10.10}
\\
RCC (ours) &\textbf{0.445}&\textbf{3.63}&\textbf{0.549}&\textbf{5.13}&\textbf{0.484}&\textbf{5.07}&0.530&11.90
\\

% Logit-based &00.00$\pm$00.00&00.00$\pm$00.00&00.00$\pm$00.00&00.00$\pm$00.00&00.00$\pm$00.00&00.00$\pm$00.00&00.00$\pm$00.00&00.00$\pm$00.00
% \\
% \midrule
% k=10 ( & & & & \\
% \midrule
% Cosine similarity \\
% RCC($\alpha$=0.0) (ours) & 52.27 & 15.84
% \\
% RCC($\alpha$=0.1) (ours) & 66.19 & 10.48
% \\
% RCC($\alpha$=0.2) (ours) & 68.90 & 5.55
% \\
% RCC($\alpha$=0.3) (ours) & 71.13 & \textbf{5.31}
% \\
% RCC($\alpha$=0.4) (ours) & 73.62 & 9.56
% \\
% RCC($\alpha$=0.5) (ours) & 74.53 & 12.75
% \\
% RCC($\alpha$=0.6) (ours) & 74.48 & 18.30
% \\
% RCC($\alpha$=0.7) (ours) & 72.92 & 22.23
% \\
% RCC($\alpha$=0.8) (ours) & 66.89 & 26.38
% \\
% RCC($\alpha$=0.9) (ours) & 57.93 & 30.14
% \\
% RCC($\alpha$=1.0) (ours) & 48.36 & 40.34
% \\
% \midrule
% Attention\\
% RCC (ours)\\
\bottomrule
\end{tabular}
\end{table*}

\section{Proposed Method}
\label{sec:method}

\textbf{Problem Definition:}
Given an instruction sequence $\mathbf{x}_{in}$, a large language model $\mathcal{M}$ generates a response $\mathbf{x}_{out}$ composed of a token sequence $\mathbf{x}_{out}=[x_{out}^1, x_{out}^2,...,x_{out}^{L_{out}}]$ with length $L_{out}$. The response $\mathbf{x}_{out}$ contains a reasoning sequence and final answer: $\mathbf{x}_{out} = [\mathbf{x}_{cot},\mathbf{x}_{ans}]$. The reasoning sequence $\mathbf{x}_{cot}=[\mathbf{s}_1, \mathbf{s}_2,...,\mathbf{s}_{L_{cot}}]$ can be naturally split into reasoning steps explicitly with clear "step-by-step" text or implicitly, like a sentence-based logical reasoning chain, where each reasoning step is also a token sequence: $\mathbf{s}_i = [x_{s_i}^1, x_{s_i}^2,...,x_{s_i}^{L_{s_i}}]$.

For each token $x$ in the generated response $\mathbf{x}_{out}$, the model $\mathcal{M}$ produces a logit vector $\mathbf{u}$ where higher logit values indicate higher probability that this token would appear in the sequence. After a softmax operation on the logit vector, we obtain $\mathbf{u'}_{out}^{i}$, and the $j$-th value in the vector represents the probability of the $j$-th token appearing in the sequence:
$$\mathcal{P}_\mathcal{M}(x_{out}^{i}|x_{out}^{i-1},x_{out}^{i-2},...,x_{out}^{1}, \mathbf{x}_{in})$$

The length of the logit vector $k$ is usually less than the whole vocabulary, and the top-$k$ strategy is commonly adopted during inference where only the highest $k$ logits are retained, since most tokens contribute negligible probability mass~\cite{meister-etal-2023-efficacy}. Our target is to quantify the uncertainty by estimating the confidence of the answer $\mathbf{x}_{ans}$ given the instruction $\mathbf{x}_{in}$:
$$\mathcal{P}_\mathcal{M}(\mathbf{x}_{ans}|\mathbf{x}_{in})$$

%nadeem-etal-2020-systematic,

However, using the joint probability $\mathcal{P}_\mathcal{M}(\mathbf{x}_{out}|\mathbf{x}_{in}) = \mathcal{P}_\mathcal{M}(\mathbf{x}_{ans},\mathbf{x}_{cot}|\mathbf{x}_{in})$ directly is unreliable since the reasoning process contains auxiliary tokens like transition words, polite expressions, and irrelevant explanations, leading to lower probability for longer responses~\cite{li2025language}. The ideal form of the probability should take a weighted average of all possible reasoning paths to assess the credibility of the final answer, but the number of all paths is exponential (vocabulary size to the power of sequence length), making exact computation intractable. Therefore, our problem is to construct an approximate probability $\hat{\mathcal{P}}_\mathcal{M} \approx \mathcal{P}_\mathcal{M}(\mathbf{x}_{ans}|\mathbf{x}_{in})$.

\noindent
\textbf{Correlation Chain:} As illustrated in Figure~\ref{fig:overview}, we consider the whole inference process as a sequence of signals composed of the instruction, reasoning, and final answer: $[\mathbf{x}_{in},\mathbf{x}_{cot},\mathbf{x}_{ans}]$. The reasoning part can be split into a detailed chain: $[\mathbf{x}_{in},\mathbf{s}_1,...,\mathbf{s}_{L_{cot}},\mathbf{x}_{ans}]$. Due to the existence of numerous tokens unrelated to the main task in the instruction $\mathbf{x}_{in}$, we filter these unrelated tokens by removing those with low attention weights. We first build an attention matrix between consecutive reasoning steps to calculate the correlation intensity between tokens.

\begin{definition}[Attention Matrix]
An \emph{attention matrix} $\mathbf{A}_i$ has shape $L_{s_i} \times L_{s_{i+1}}$, where each row $\mathbf{v}_j$ represents the attention between the $j$-th token in $\mathbf{s}_i$ and all tokens in $\mathbf{s}_{i+1}$:
$$ \mathbf{v}_j = \frac{\mathbf{x}_{s_i}^j \cdot \mathbf{s}_{i+1}^T}{\sqrt{d}} $$
where $d$ is the embedding dimension.
\end{definition}

The first attention matrix $\mathbf{A}_0$ is initialized with the instruction, where we consider $\mathbf{s}_0 = \mathbf{x}_{in}$. For each value $v_j^k$ in the attention matrix row $\mathbf{v}_j$, we apply softmax normalization to obtain normalized attention weights in $[0,1]$, forming the normalized matrix $\mathbf{A}'$.
% $$ v_j^{'k} = \frac{e^{v_j^{k}}}{\sum_{m=1}^{L_{s_{i+1}}} e^{v_j^{m}}}$$
With a threshold $\mu$, we filter low attention items in the matrix, which are considered as irrelevant tokens, using the Heaviside step function $w^{k}_{j} = \mathcal{H}(v^{'k}_{j}-\mu)$ for each element $v'_{jk}$ in the matrix, forming the filtered attention matrix $\mathbf{W}_i$.

\noindent
\textbf{Temporal Hidden Confidence:}
The softmaxed logit vectors in each reasoning step $\mathbf{s}_i$ form a confidence chain $\mathbf{c}_i = [c_i^1, c_i^2,...,c_i^{L_{s_i}}]$ where each element represents the probability of the corresponding token in $\mathbf{s}_i$. As shown in Figure~\ref{fig:overview}, the correlated confidence for step $\mathbf{s}_i$ is computed by weighted averaging the confidence values based on attention connections from the previous step. For every element $r_j$ in the vector $\mathbf{W}_{i-1} \mathbf{c}_i^{T}$, the correlated confidence is:
$$q_i = \frac{\sum_{j=1}^{L_{s_{i-1}}} r_j \cdot \mathbf{1}_{\{r_j \neq 0\}}}
         {\sum_{j=1}^{L_{s_{i-1}}} \mathbf{1}_{\{r_j \neq 0\}}}$$

The accumulated confidence $p$ propagates through each step with a recurrent update:
$p_i = \delta q_i + (1-\delta)p_{i-1}$
where $\delta \in (0,1)$ is a hyperparameter controlling the balance between current and historical confidence. For initialization, $p_1 = q_1$. The final uncertainty estimate is $\hat{\mathcal{P}}_\mathcal{M}(\mathbf{x}_{ans}|\mathbf{x}_{in}) = p_{n}$ where n is the total amount of reasoning steps.

\section{Experiment}
\label{sec:exp}

% \subsection{Experimental Setup}

\noindent
\textbf{Experimental Setup:} We employ two challenging benchmarks to evaluate our framework.  For our study, we utilize the mathematics subsets of the {GAOKAO Benchmark}~\cite{zhang2023evaluating}, which offers a suite of complex multi-step reasoning problems. The {CLadder} dataset~\cite{jin2023cladder} is specifically designed to test causal reasoning, categorizing into three levels of causal query complexity: associational, interventional, and counterfactual. This allows for a granular analysis of uncertainty quantification across different forms of logical deduction.
We test our method on two leading open-source LLMs known for their robust reasoning abilities and availability for in-depth analysis: {Qwen3-8B}\footnote{https://huggingface.co/Qwen/Qwen3-8B} and {Gemma3-12B}\footnote{https://huggingface.co/google/gemma-3-12b-it}.
Our experiments were conducted on a cluster of NVIDIA B200 GPUs. Inference settings were consistent across all tests:  a temperature of $\tau=0$ for deterministic output, top-$k$ sampling with $k=50$ to manage the vocabulary and the attention threshold $\mu$ is 0.5 as ~\cite{li2025language} suggested. We systematically explored different values for the propagation weight $\delta$ to evaluate their effect on the model's performance and calibration.

\begin{figure}[h!]
  \centering
  \includegraphics[width=0.48\textwidth]{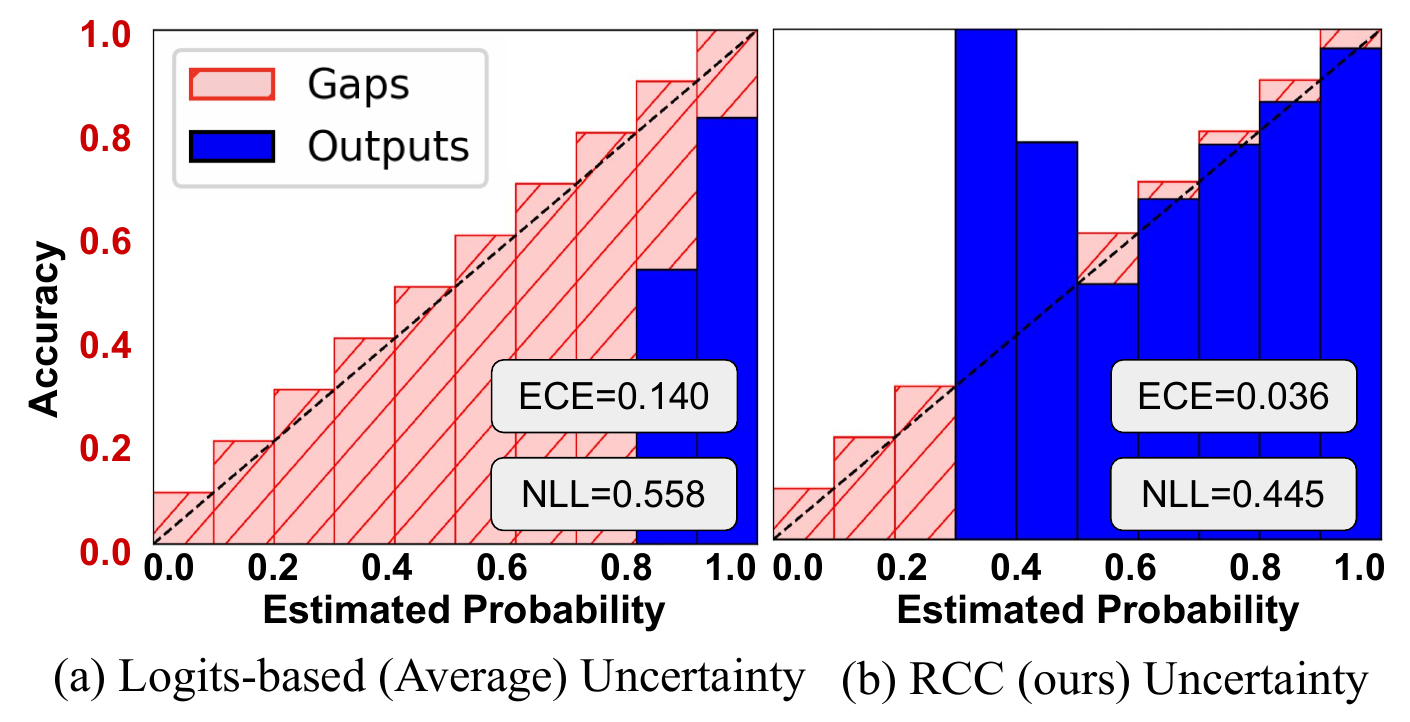}
  \caption{Reliability diagrams on the GAOKAO MATH dataset, using the Qwen3-8B model. }
  \label{fig:reliability_diagrams}
\end{figure}

\noindent
\textbf{Baselines and Metrics:} To contextualize the performance of our framework, we compare it against a set of established baselines. These include direct confidence measures derived from the language model's output: average logits and the logits score on the final answer. We also benchmark our method against two prominent uncertainty quantification techniques: Self-Consistency~\cite{wang2023selfconsistency}, and Verbalized Uncertainty~\cite{xiong2024can}. Furthermore, we include two recent, state-of-the-art approaches:  the Shifting Attention to Relevance technique~\cite{duan2024shifting} and the Attention Chain (UQAC) method~\cite{li2025language}, to provide a comprehensive comparison.
Our evaluation focuses on both the discriminative power and calibration of our model's uncertainty estimates. We use Negative Log-Likelihood (NLL) to assess the sharpness and quality of the predicted probabilities. Recognizing that a low NLL does not guarantee well-calibrated confidence scores, we also place significant emphasis on Expected Calibration Error (ECE). This metric quantifies the divergence between predicted confidence and actual accuracy, providing a more reliable measure of the model's trustworthiness. Additionally, we use calibration plots to visually inspect how well the predicted confidence scores align with empirical accuracy across different confidence bins.

% We compare RCC against four established uncertainty quantification methods:
% \begin{itemize}
% \item \textbf{Token Probability (TP)}: Baseline approach computing joint probability through direct token probability multiplication, representing the standard maximum likelihood estimation.
% \item \textbf{Self-Consistency (SC)}: Samples 5 reasoning paths with temperature 0.7 and measures answer consistency, following the majority voting principle.
% \item \textbf{Verbalized Confidence (VC)}~\cite{tian2023just}: Prompts models to explicitly state confidence levels (0-100\%) after generating answers.
% \item \textbf{Attention-based UQ (AUQ)}~\cite{vazhentsev2025uncertainty}: Recent attention-based method using uncertainty-aware heads without temporal modeling.
% \end{itemize}

% Evaluation employs three complementary metrics: Expected Calibration Error (ECE) with 10 bins measuring confidence-accuracy alignment, AUROC evaluating discrimination between correct/incorrect predictions, and Brier Score assessing overall probabilistic prediction quality.

\section{Results and Discussion}
\label{sec:result}

\noindent
\textbf{Comparison of Results:} Our main findings, summarized in Table~\ref{tab:perf}, show that the RCC framework has a significant advantage in calibration across the GAOKO and CLadder datasets, consistently achieving a lower Expected Calibration Error (ECE) compared to baseline methods. This superior calibration is visually confirmed in our reliability diagrams (Figure~\ref{fig:reliability_diagrams}), where RCC's outputs align closely with the ideal diagonal line. While common baselines like average logits tend to produce uncalibrated scores and others such as Self-Consistency may perform well on a single metric, our framework strikes a superior balance, providing both low Negative Log-Likelihood (NLL) and exceptional calibration.

\begin{figure}[h!]
  \centering
  \includegraphics[width=0.48\textwidth]{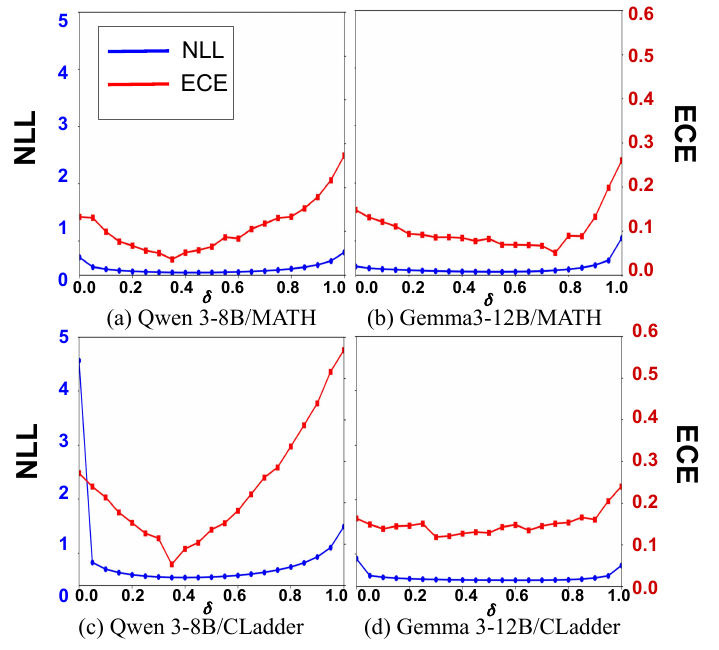}
  \caption{Performance of the RCC model with varying propagation weight $\delta$ on different models and datasets.}
  \label{fig:hyper}
\end{figure}

% Our main findings are summarized in Table ~\ref{tab:perf}, which details the performance of our framework and the baselines on the GAOKAO and CLadder datasets. Our approach demonstrates a significant advantage in calibration, consistently achieving lower Expected Calibration Error (ECE) compared to other methods. This is visually confirmed in our Reliability Diagrams (Figure 2), where our method's calibration curve aligns closely with the ideal diagonal line, indicating that its predicted confidence scores accurately reflect its true performance.

% In contrast, methods like average logits and final logits tend to produce uncalibrated scores, where high confidence does not reliably correlate with accuracy. While some baselines, such as Self-Consistency and~\cite{wang2023selfconsistency} others~\cite{xiong2024can}, show competitive performance on one metric, our framework strikes a superior balance, providing both low Negative Log-Likelihood (NLL) and exceptional calibration. Verbalized Uncertainty performs poorly across the board, suggesting that it is not well-suited for these complex reasoning tasks without extensive domain-specific fine-tuning.

\noindent\textbf{Hyperparameter Analysis:}
Our analysis of the propagation weight $\delta$, illustrated in Figure~\ref{fig:hyper}, reveals its critical impact on performance. This hyperparameter balances the influence between the current step's confidence (weighted by $\delta$) and the accumulated historical confidence (weighted by $1-\delta$). The plots show a clear trade-off: a high $\delta$ gives more weight to the current step's confidence, which can aggressively sharpen the probability distribution (leading to a lower NLL) but may cause overconfidence and harm calibration (increasing ECE) by ignoring a history of uncertainty. Conversely, a low $\delta$ relies too heavily on past steps and may be overly conservative. The optimal settings, typically found in the mid-range of $\delta \in [0.2, 0.6]$, are those that balance this trade-off, as a reliable model must be both sharp and well-calibrated. These results highlight the importance of carefully tuning this parameter to achieve robust performance.

\noindent
\textbf{Discussion:} Our empirical results demonstrate that the RCC framework effectively bridges the gap between model accuracy and calibration, generating sharp and reliable confidence scores where traditional baselines like average logits often fail. By propagating uncertainty through the model's internal reasoning process, our method achieves low ECE values, indicating that its confidence is highly correlated with actual performance. A key challenge for this framework, presenting an avenue for future research, lies in handling the complex multi-headed attention of modern LLMs. Since different heads serve distinct functions, dynamically determining which ones to leverage for uncertainty propagation is an open problem. This observation highlights a fundamental trade-off in uncertainty quantification: aggressive propagation can sharpen confidence scores but may push the model toward overconfidence, thereby harming calibration.

% Our empirical results demonstrate that the RCC framework effectively bridges the critical gap between model accuracy and calibration. While traditional baselines like average and final logits provide a rough measure of confidence, they often yield uncalibrated scores that fail to reflect the model's true reliability. Our method, by propagating uncertainty through the model's internal reasoning process, generates confidence scores that are not only sharp but also highly correlated with actual performance, as evidenced by our low ECE values.

% A key challenge for this framework lies in its handling of the complex, multi-headed attention mechanism of modern LLMs. While we've shown the effectiveness of an attention chain-based approach, a single model's multiple attention heads often serve distinct functions—for example, focusing on word semantics, syntactic structures, or coreferential relationships. Dynamically determining which attention heads to leverage for uncertainty propagation remains an open problem. A one-size-fits-all approach may not fully capture the nuanced contributions of each head, presenting a key area for future research. This observation further underscores a fundamental trade-off in uncertainty quantification: while aggressive propagation can sharpen confidence scores, leading to a lower NLL, it may also push the model toward overconfidence, thereby increasing ECE.

\section{Conclusion}
\label{sec:conclusion}
In this work, we introduced the Recurrent Confidence Chain, a novel framework to quantify uncertainty in LLMs. By leveraging inter-step attention for semantic correlation and a recurrent mechanism to propagate confidence history, our method provides better estimates for complex reasoning tasks. The results show our approach outperforms existing baselines by achieving a superior balance between predictive performance and calibration, offering a more trustworthy measure of model certainty for high-stakes applications.

\newpage
\bibliographystyle{IEEEbib}
\bibliography{strings,refs}

@article{korbak2025chain,
  title={Chain of thought monitorability: A new and fragile opportunity for ai safety},
  author={Korbak, Tomek and Balesni, Mikita and Barnes, Elizabeth and Bengio, Yoshua and Benton, Joe and Bloom, Joseph and Chen, Mark and Cooney, Alan and Dafoe, Allan and Dragan, Anca and others},
  journal={arXiv:2507.11473},
  year={2025}
}

@article{duan2025uprop,
  title={UProp: Investigating the Uncertainty Propagation of LLMs in Multi-Step Agentic Decision-Making},
  author={Duan, Jinhao and Diffenderfer, James and Madireddy, Sandeep and Chen, Tianlong and Kailkhura, Bhavya and Xu, Kaidi},
  journal={arXiv:2506.17419},
  year={2025}
}

@article{he2025mmboundary,
  title={Mmboundary: Advancing mllm knowledge boundary awareness through reasoning step confidence calibration},
  author={He, Zhitao and Polisetty, Sandeep and Fan, Zhiyuan and Huang, Yuchen and Wu, Shujin and Fung, Yi R},
  journal={arXiv:2505.23224},
  year={2025}
}

@inproceedings{mao-etal-2025-temporalizing,
    title = "Temporalizing Confidence: Evaluation of Chain-of-Thought Reasoning with Signal Temporal Logic",
    author = "Mao, Zhenjiang  and
      Bisliouk, Artem  and
      Nama, Rohith  and
      Ruchkin, Ivan",
    booktitle = "Proceedings of the 20th Workshop on Innovative Use of NLP for Building Educational Applications (BEA 2025)",
    month = jul,
    year = "2025",
    address = "Vienna, Austria",
    publisher = "Association for Computational Linguistics",
    url = "https://aclanthology.org/2025.bea-1.65/",
    doi = "10.18653/v1/2025.bea-1.65",
    pages = "882--890",
    ISBN = "979-8-89176-270-1"
}

@inproceedings{
azaria2023the,
title={The Internal State of an {LLM} Knows When It's Lying},
author={Amos Azaria and Tom Mitchell},
booktitle={Conference on Empirical Methods in Natural Language Processing},
year={2023},
url={https://openreview.net/forum?id=y2V6YgLaW7}
}

@inproceedings{zhang-etal-2024-luq,
    title = "{LUQ}: Long-text Uncertainty Quantification for {LLM}s",
    author = "Zhang, Caiqi  and
      Liu, Fangyu  and
      Basaldella, Marco  and
      Collier, Nigel",
    booktitle = "Proceedings of the 2024 Conference on Empirical Methods in Natural Language Processing",
    month = nov,
    year = "2024",
    address = "Miami, Florida, USA",
    publisher = "Association for Computational Linguistics",
    url = "https://aclanthology.org/2024.emnlp-main.299/",
    doi = "10.18653/v1/2024.emnlp-main.299",
    pages = "5244--5262"
}

@inproceedings{
da2025understanding,
title={Understanding the Uncertainty of {LLM} Explanations: A Perspective Based on Reasoning Topology},
author={Longchao Da and Xiaoou Liu and Jiaxin Dai and Lu Cheng and Yaqing Wang and Hua Wei},
booktitle={Second Conference on Language Modeling},
year={2025},
url={https://openreview.net/forum?id=p4wZfBFgyI}
}

@misc{
duan2024shifting,
title={Shifting Attention to Relevance: Towards the Uncertainty Estimation of Large Language Models},
author={Jinhao Duan and Hao Cheng and Shiqi Wang and Alex Zavalny and Chenan Wang and Renjing Xu and Bhavya Kailkhura and Kaidi Xu},
year={2024},
url={https://openreview.net/forum?id=yZJapMWdHZ}
}

@inproceedings{
wang2023selfconsistency,
title={Self-Consistency Improves Chain of Thought Reasoning in Language Models},
author={Xuezhi Wang and Jason Wei and Dale Schuurmans and Quoc V Le and Ed H. Chi and Sharan Narang and Aakanksha Chowdhery and Denny Zhou},
booktitle={The Eleventh International Conference on Learning Representations },
year={2023},
url={https://openreview.net/forum?id=1PL1NIMMrw}
}

@inproceedings{
xiong2024can,
title={Can {LLM}s Express Their Uncertainty? An Empirical Evaluation of Confidence Elicitation in {LLM}s},
author={Miao Xiong and Zhiyuan Hu and Xinyang Lu and YIFEI LI and Jie Fu and Junxian He and Bryan Hooi},
booktitle={The Twelfth International Conference on Learning Representations},
year={2024},
url={https://openreview.net/forum?id=gjeQKFxFpZ}
}

@inproceedings{gao-etal-2024-spuq,
    title = "{SPUQ}: Perturbation-Based Uncertainty Quantification for Large Language Models",
    author = "Gao, Xiang  and
      Zhang, Jiaxin  and
      Mouatadid, Lalla  and
      Das, Kamalika",
    booktitle = "Proceedings of the 18th Conference of the European Chapter of the Association for Computational Linguistics",
    month = mar,
    year = "2024",
    publisher = "Association for Computational Linguistics",
    url = "https://aclanthology.org/2024.eacl-long.143/",
    doi = "10.18653/v1/2024.eacl-long.143"
}

@inproceedings{jin2023cladder,
    author = {Zhijing Jin and Yuen Chen and Felix Leeb and Luigi Gresele and Ojasv Kamal and Zhiheng Lyu and Kevin Blin and Fernando Gonzalez and Max Kleiman-Weiner and Mrinmaya Sachan and Bernhard Sch{\"{o}}lkopf},
    title = "{CL}adder: {A}ssessing Causal Reasoning in Language Models",
    year = "2023",
    booktitle = "NeurIPS",
    url = "https://openreview.net/forum?id=e2wtjx0Yqu",
}

@article{zhang2023evaluating,
  title={Evaluating the performance of large language models on gaokao benchmark},
  author={Zhang, Xiaotian and Li, Chunyang and Zong, Yi and Ying, Zhengyu and He, Liang and Qiu, Xipeng},
  journal={arXiv:2305.12474},
  year={2023}
}

@inproceedings{tian-etal-2023-just,
    title = "Just Ask for Calibration: Strategies for Eliciting Calibrated Confidence Scores from Language Models Fine-Tuned with Human Feedback",
    author = "Tian, Katherine  and
      Mitchell, Eric  and
      Zhou, Allan  and
      Sharma, Archit  and
      Rafailov, Rafael  and
      Yao, Huaxiu  and
      Finn, Chelsea  and
      Manning, Christopher",
    booktitle = "Proceedings of the Conference on Empirical Methods in Natural Language Processing",
    month = dec,
    year = "2023",
    address = "Singapore",
    publisher = "Association for Computational Linguistics",
    url = "https://aclanthology.org/2023.emnlp-main.330/",
    doi = "10.18653/v1/2023.emnlp-main.330"
}

@inproceedings{
li2025language,
title={Language Model Uncertainty Quantification with Attention Chain},
author={Yinghao Li and Rushi Qiang and Lama Moukheiber and Chao Zhang},
booktitle={Second Conference on Language Modeling},
year={2025},
url={https://openreview.net/forum?id=QTrW2HWNXe}
}

@inproceedings{meister-etal-2023-efficacy,
    title = "On the Efficacy of Sampling Adapters",
    author = "Meister, Clara  and
      Pimentel, Tiago  and
      Malagutti, Luca  and
      Wilcox, Ethan  and
      Cotterell, Ryan",
    booktitle = "Proceedings of the 61st Annual Meeting of the Association for Computational Linguistics",
    month = jul,
    year = "2023",
    address = "Toronto, Canada",
    publisher = "Association for Computational Linguistics",
    url = "https://aclanthology.org/2023.acl-long.80/",
    doi = "10.18653/v1/2023.acl-long.80",
    pages = "1437--1455"
}

@inproceedings{
liu2025adaptivestep,
title={AdaptiveStep: Automatically Dividing Reasoning Step through Model Confidence},
author={Yuliang Liu and Junjie Lu and Chaofeng Qu and Zhaoling Chen and Zefan Cai and Jason Klein Liu and Chonghan Liu and Yunhui Xia and Li Zhao and Jiang Bian and Chuheng Zhang and Wei Shen and Zhouhan Lin},
booktitle={42nd International Conference on Machine Learning},
year={2025},
url={https://openreview.net/forum?id=ViRFgwVjk0}
}

@article{shorinwa2025survey,
  title={A survey on uncertainty quantification of large language models: Taxonomy, open research challenges, and future directions},
  author={Shorinwa, Ola and Mei, Zhiting and Lidard, Justin and Ren, Allen Z and Majumdar, Anirudha},
  journal={ACM Computing Surveys},
  year={2025},
  publisher={ACM New York, NY}
}

\end{document}